\documentclass[12pt,a4paper]{article}
\usepackage{epsfig}

\input{tcilatex}

\begin{document}

\title{Modelling Complexity in Musical Rhythm}
\author{ Cheng-Yuan Liou$^{1}$, Tai-Hei Wu, and Chia-Ying Lee \\
{\small Department of Computer Science and Information Engineering }\\
{\small National Taiwan University }\\
{\small Supported by National Science Council }}
\maketitle

\begin{abstract}
This paper constructs a tree structure for the music rhythm using the
L-system. It models the structure as an automata and derives its complexity.
It also solves the complexity for the L-system. This complexity can resolve
the similarity between trees. This complexity serves as a measure of
psychological complexity for rhythms. It resolves the music complexity of
various compositions including the Mozart effect K488.


L-system, automata, temporal associative memory, inverse

problem, rewriting rule, bracketed string, tree similarity

the number of text pages of the manuscript 21

the number of figures 13

the number of tables \ \ 2

$^{1}$Correspondence: Department of Computer Science and

Information Engineering, National Taiwan University, Taipei,

Taiwan, 106, R.O.C., Tel: 8862 23625336 ext 515,

Fax: 8862 23628167, email: cyliou@csie.ntu.edu.tw
\end{abstract}

\section[Introduction]{Introduction}

This paper presents a computational model to interpret the 'Mozart effect'
K488. This effect has been discussed extensively and seriously among both
psychology and music perception societies using various experimental
techniques [Rauscher et al. 1993]. Instead of experiments, this paper
constructs a computation model to resolve the effect. This model starts with
a tree structure for quantized rhythm beats based on the theory by
Loguet-Higgins [Longuet-Higgins, 1987]. The tree is then modeled as an
automata and its complexity is derived by way of the L-system [Prusinkiewicz
and Lindenmayer 1990, Prusinkiewicz 1986]. The quantization tree will be
briefly introduced in this section. The L-system will be introduced in the
next section. \ The automata rewriting rule associated with the L-system
will be included also in the next section. The similarity between trees by
way of rewriting rules is defined in Section 3. The tree complexity is
derived in Section 4. This complexity serves as a measure for the perception
of musical rhythms [Desain and Windsor 2000; Yeston 1976] and resolves the
effect.

The key features of music perception and composition are rhythm, melody and
harmony. Rhythm is formed through the alternation of long and short notes,
or through repetition of strong and weak dynamics [Cooper and Meyer 1960].
Because one metrical unit, such as a measure or a half note, can often be
divided into two or three sub-units (illustrated in Figures \ref{figure 1}
and \ref{figure 2}), this rhythm is endowed with a clear hierarchical
structure [Longuet-Higgins, 1987; Lerdahl and Jackendoff 1983]. Note that
Longuet Higgins grammar for rhythm is different from his musical parser for
handeling performances, which is far more sophisticated than time-grid
round-off. To represent the hierarchical characteristics of rhythm, we need
to seek a system that possesses such a nature. Fortunately, the plant
kingdom is rich with branching structures, in which branches are derived
from roots. In fact, the structure shown in Figure \ref{figure 1} is that of
a binary tree. L-systems (Lindenmayer systems) [Prusinkiewicz and
Lindenmayer 1990, Prusinkiewicz 1986] are designed to model plant
development, see [McCormack 1993]. Therefore, it is natural to construct the
rhythm representation by using an L-system [Prusinkiewicz 1986; Worth and
Stepney 2005]. A background of L-Systems applied to art and music is in the
website in Reference.

We will show how such a tree structure and its related parts can be
constructed. We expect that the L-system can capture the rhythm nature. We
now review the music tree by Loguet-Higgins.

Figure \ref{figure 1} shows that in each level of a tree a half note is
represented by a different metrical unit. In the highest level, the metrical
unit is a half note; in the next level, the unit is a quarter note; in the
lowest level, the unit is an eighth note. The total duration in each level
is equal to a half note. This structure can be extended to a measure or
more, a composition as in Figure \ref{figure 2}.

\FRAME{fhFU}{0.9279in}{1.0334in}{0pt}{\Qcb{ One metrical can be divided into
two or three subunits.}}{\Qlb{figure 1}}{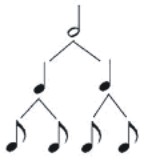}{\special{language
"Scientific Word";type "GRAPHIC";maintain-aspect-ratio TRUE;display
"USEDEF";valid_file "F";width 0.9279in;height 1.0334in;depth
0pt;original-width 1.4996in;original-height 1.6769in;cropleft "0";croptop
"1";cropright "1";cropbottom "0";filename 'fig1.jpg';file-properties
"XNPEU";}}\FRAME{fhFU}{2.9594in}{1.235in}{0pt}{\Qcb{Musical tree of
Beethoven's Piano Sonata No. 6, Mov. 3.}}{\Qlb{figure 2}}{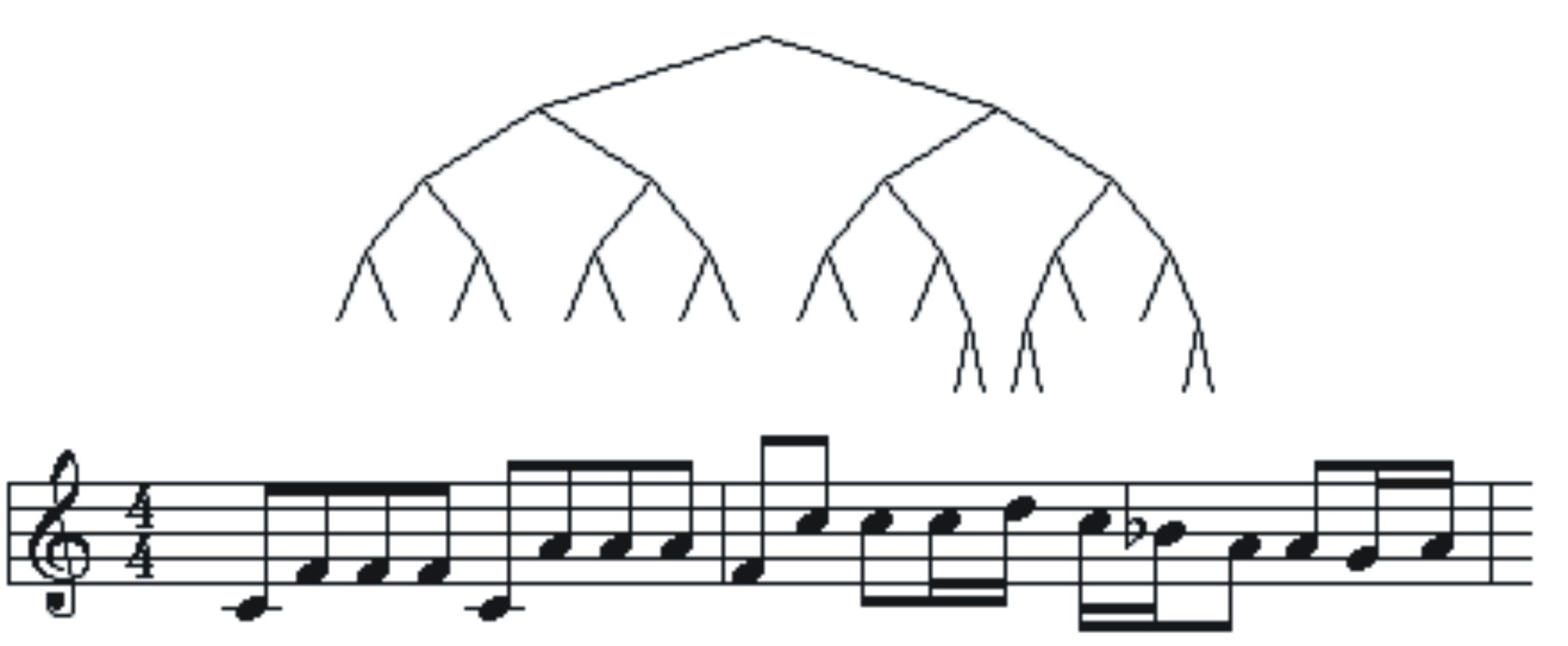}{\special%
{language "Scientific Word";type "GRAPHIC";maintain-aspect-ratio
TRUE;display "USEDEF";valid_file "F";width 2.9594in;height 1.235in;depth
0pt;original-width 6.0935in;original-height 2.5105in;cropleft "0";croptop
"1";cropright "1";cropbottom "0";filename 'fig2.jpg';file-properties
"XNPEU";}}

H. C. Longuet-Higgins introduced this kind of tree in the 1970s. The tree
has been extensively studied both experimently and theoretically. Note that
there are many other theories [Cooper and Meyer 1960; Yeston 1976; Lerdahl
and Jackendoff 1983; Desain and Windsor 2000]. Since we prefer a
computational approach to resolve the effect, we will not use their
theories. In order to give computers the musicianship necessary to
transcribe a melody into a score, he used tree structures to represent
rhythmic groupings. In his theory of music perception, the essential task in
perceiving the rhythmic structure of a melody is to identify the time of
occurrence of each beat. Therefore, his theory can be applied to Western
music with regular beats. In Western music, the most common subdivisions of
each beat are into two or three shorter metrical units, and these shorter
metrical units can be further subdivided into two or three units. Tracking
from the start of a melody, when a beat or a fraction of a beat is
interrupted by the onset of a note, it is divided into shorter metrical
units. After this process of division, every note will find itself at the
beginning of an uninterrupted metrical unit. The metrical units can be
considered as the nodes of a tree in which each non-terminal node has two or
three descendants. The terminal nodes for a beat are the shortest metrical
units that the beat is divided into. Every terminal node in the tree will
eventually be attached either to a rest or to a note sounded or tied. It is
natural to include and elaborate rests in the tree model as those done in
[Longuet-Higgins, 1987].

We will employ the perception factors discussed by Longuet-Higgins, such as
tolerance, syncopation, rhythmic ambiguity, regular passages, to construct
L-systems for rhythms.

\section[Rhythm represented with tree structure]{Rhythm represented with
tree structure}

A rhythmic tree as described above is a tree of which each subtree is also a
rhythmic tree. Each tree node has two or three children (branches or
descendants). Each node in the tree represents the total beat duration that
is equal to the sum of all those of its descendants. The root node has a
duration length that is equal to the length of the whole note sequence.

Note that when we attempt to split a note sequence into two subsequences
with equal duration lengths, we usually obtain two unequal length
subsequences. This is because a note connecting the two subsequences has
been split into two submetrical units. The preceding portion belongs to the
preceding subsequences and the later portion belongs to the later
subsequence. We will mark those units to identify their subsequences.

These two subsequences represent two different subtrees of the root node. We
further divide each subsequence into sub-subsequences, which are also
rhythmic trees, and so on. This dividing process is completed when a tree
node contains a single note. This single note may possibly be the one which
has been split into two portions. This is in some sense similar to an
algorithm for note quantization and is a standard practice in MIDI rendering
of music. In practice we will quantize notes using the finest note among
dotted notes (e.g., 1/4, 1/8, 1/16, dotted notes, etc.) without loosing most
of the interesting details. We plot two such trees in Figures \ref{figure 2}
and \ref{figure 3}. The notes shown in Figure \ref{figure 3} are part of the
whole tree of the beginning of Rachnaminoff's Piano Concerto No.3, Movement
1. The two notes in the rectangle have been split using our rhythmic tree
process.

\FRAME{fhFU}{3.1799in}{2.0539in}{0pt}{\Qcb{Musical tree of Rachmaninoff's
Piano Concerto No.3, Mov. 1.}}{\Qlb{figure 3}}{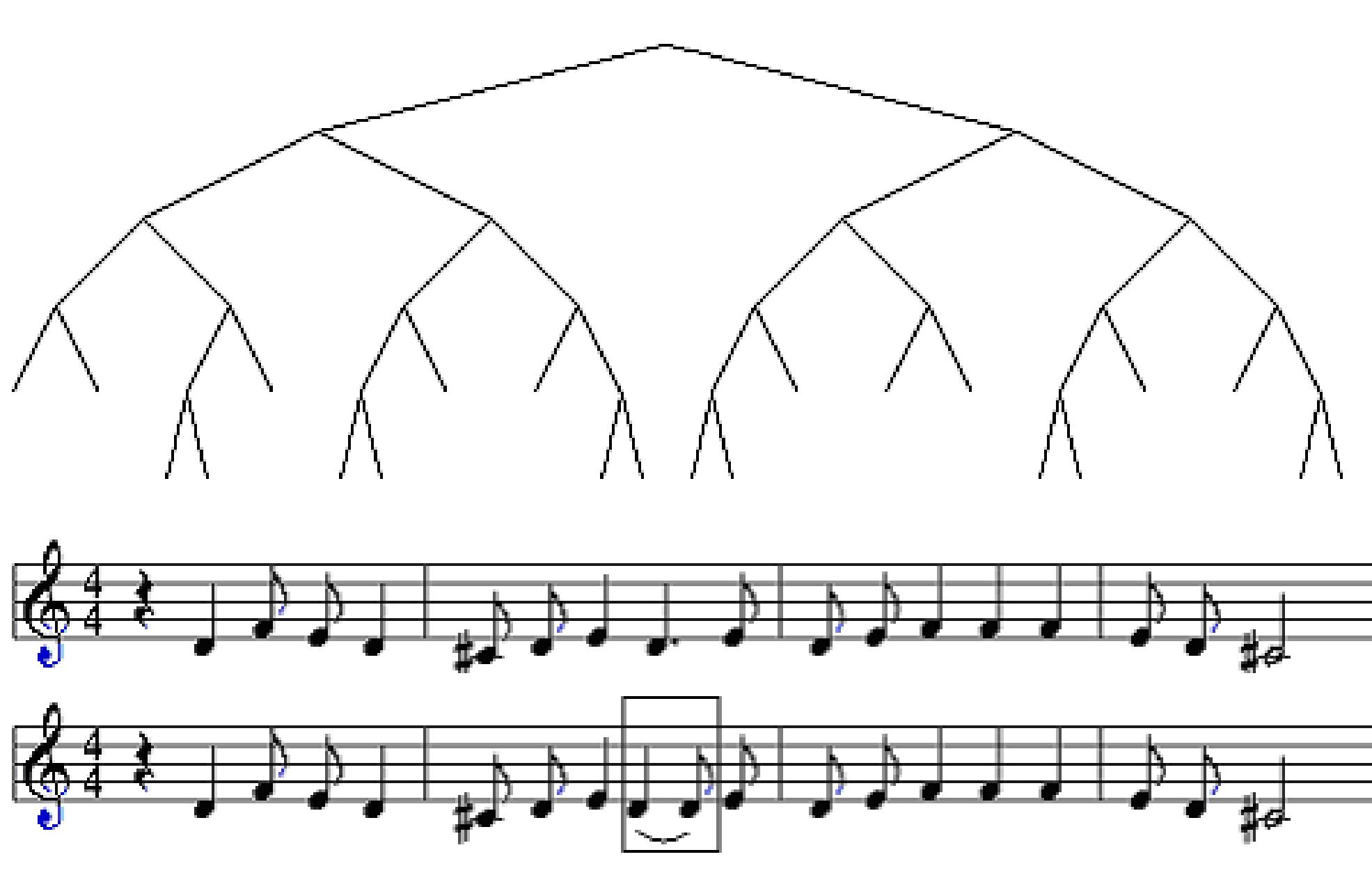}{\special{language
"Scientific Word";type "GRAPHIC";maintain-aspect-ratio TRUE;display
"USEDEF";valid_file "F";width 3.1799in;height 2.0539in;depth
0pt;original-width 6.57in;original-height 4.2229in;cropleft "0";croptop
"1";cropright "1";cropbottom "0";filename 'fig3.jpg';file-properties
"XNPEU";}}

\subsection[Rhythm represented with L-systems]{Rhythm represented with
L-system}

To express the hierarchical characteristics of rhythm, we need a data
structure that possesses such a hierarchical nature. Fortunately, the plant
kingdom is dominated by branching structures, in which branches are derived
from roots. L-systems are designed to model plant development. Therefore, it
is practicable to construct a rhythmic representation by using L-systems.

The Lindenmayer system, or L-system for short, was introduced by the
biologist Aristid Lindenmayer in 1968 [Lindenmayer 1968]. It was conceived
as a mathematical theory of plant development. The central concept of the
L-system is rewriting. In general, rewriting is a technique used to define
complex objects by successively replacing parts of a simple initial object,
using a set of rewriting rules or productions.

The L-system is a new type of string-rewriting mechanism. The essential
difference between Chomsky grammars and L-systems lies in the technique used
to apply productions. In Chomsky grammars, productions are applied
sequentially, whereas in L-systems, they are applied in parallel and
simultaneously to replace all the letters in a given word [McCormack 1993].

This difference reflects the biological motivation of L-systems. Productions
are intended to capture cell divisions in multi-cellular organisms, where
many divisions may occur at the same time. Moreover, there are languages
which can be generated by context-free L-systems but not by context-free
Chomsky grammars.

Here we introduce the turtle graphical interpretation of L-systems. Suppose
that there is a turtle crawling on a plane. The state of the turtle is
defined as a triplet $(x,y,\alpha )$, where the Cartesian coordinates $(x,y)$
represent the turtle's position, and the angle $\alpha $, called the
heading, is interpreted as the direction in which the turtle is facing.
Given the step size d and the angle increment $\delta $, the turtle can
respond to commands represented by the following symbols:

\begin{quote}
\begin{tabular}[t]{lp{12cm}}
F & Move forward a step of length $d$. The state of the turtle changes to $%
(x^{new},y^{new},\alpha )$, where $x^{new}=x+d\cos \alpha $ and $%
y^{new}=y+d\sin \alpha $. Draw a line segment between points $(x,y)$ and $%
(x^{new},y^{new})$. \\ 
f & Move forward a step of length $d$ without drawing a line. \\ 
+ & Turn left (counterclockwise) by angle $\delta $. The next state of the
turtle is $(x,y,\alpha +\delta )$. The positive orientation of angles is
counterclockwise. \\ 
- & Turn right (clockwise) by angle $\delta $. The next state of the turtle
is $(x,y,\alpha -\delta )$. \\ 
& 
\end{tabular}
\end{quote}

For a string composed by the above symbols, the turtle will crawl according
to commands indicated in a given order. The turtle interpretation of this
string is the figure drawn by the turtle. To draw branches of a tree, we
need two more symbols:

\begin{quote}
\begin{tabular}[t]{lp{4.73in}}
\lbrack & Push the current state of the turtle onto a pushdown stack. The
information saved on the stack contains the turtle's position and
orientation, and possibly other attributes such as the color and width of
the lines being drawn. \\ 
] & Pop a state from the stack and make it the current state of the turtle.
No line is drawn although, in general, the position of the turtle changes.
\\ 
& 
\end{tabular}
\end{quote}

These two symbols enable the turtle to go back to the root after drawing a
branch so that it can draw other branches originating from the same root.
This kind of L-system is called the bracketed L-system, and we call strings
that represent trees bracketed strings.

\subsection[Rewriting rules for rhythmic trees]{Rewriting rules for rhythmic
trees}

In music, a longer metrical unit can be replaced by the combination of
several shorter metrical units, such as a bar filled with several notes.
Given a rhythm, the direct way to represent it with an L-system is to
construct rewriting rules that replace longer metrical units with shorter
ones. Take the dotted half note for example; it can be replaced by 3 quarter
notes or by the combination of a quarter note, a dotted quarter note and an
eighth note, of a half note and a quarter note, and so on, see Figure \ref%
{figure 4}. In Figure \ref{figure 4}, 4 rewriting rules are shown that can
be used to rewrite a longer metrical unit (a dotted half note) into several
shorter metrical units (a quarter note, dotted quarter note, and so on).

\FRAME{fhFU}{1.1416in}{1.0239in}{0pt}{\Qcb{Different combinations of a
dotted half note.}}{\Qlb{figure 4}}{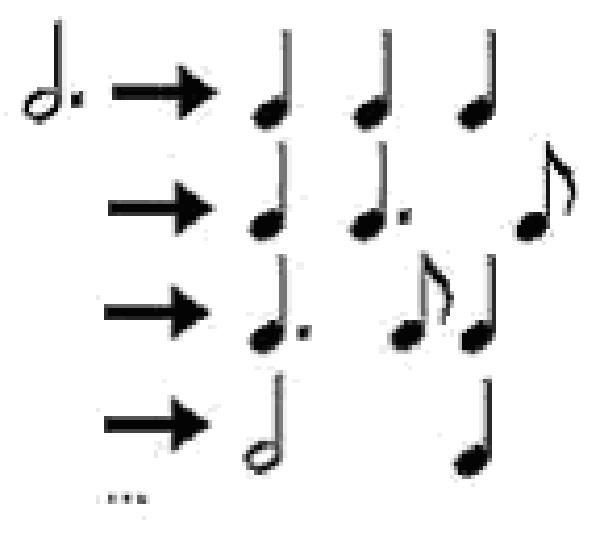}{\special{language "Scientific
Word";type "GRAPHIC";maintain-aspect-ratio TRUE;display "USEDEF";valid_file
"F";width 1.1416in;height 1.0239in;depth 0pt;original-width
1.9873in;original-height 1.7772in;cropleft "0";croptop "1";cropright
"1";cropbottom "0";filename 'fig4.jpg';file-properties "XNPEU";}}

Thus, we can regard the whole score as the longest metrical unit and
construct rewriting rules for it, recursively from the longest unit to the
shortest metrical unit. These rewriting rules can represent the rhythm
because we can regenerate the whole score by using these rules. We will not
elaborate on the rewriting rules for a score [Lee and Liou 2003]. The
rewriting rule system is formally equivalent to the bracketed L-system.

Similar to a score, a rhythmic tree can also be represented by means of
rewriting rules. After we build a rhythmic tree from a score by using the
technique described in Section 2, we may apply rewriting rules to each node
in the rhythmic tree. As a half node can be rewritten as 2 quarter nodes,
each node can be rewritten as all the subtrees attached to it. The rewriting
rules for our tree nodes always generate 2 subtrees because the technique
described in section 2 always generates 2 subtrees (children) for each node.
Note that similar qualitative results of this work can be obtained with 3 or
more subtrees. We will focus on 2 generated subtrees in this work. In Figure %
\ref{figure 5}, we represent a subtree (or a metrical unit) using the
rewriting rule \textbf{$P\rightarrow LR$}, where \textbf{$P$} denotes the
subtree we want to represent using a set of rewriting rules, \textbf{$L$}
denotes its left subtree, and \textbf{$R$} denotes its right subtree. We
call \textbf{$L$} and \textbf{$R$} the nonterminals. With this schema, we
can write the rewriting rule for the tree shown below.

\FRAME{fhFU}{3.8233in}{1.3353in}{0pt}{\Qcb{Using rewriting rules to
represent a rhythmic tree.}}{\Qlb{figure 5}}{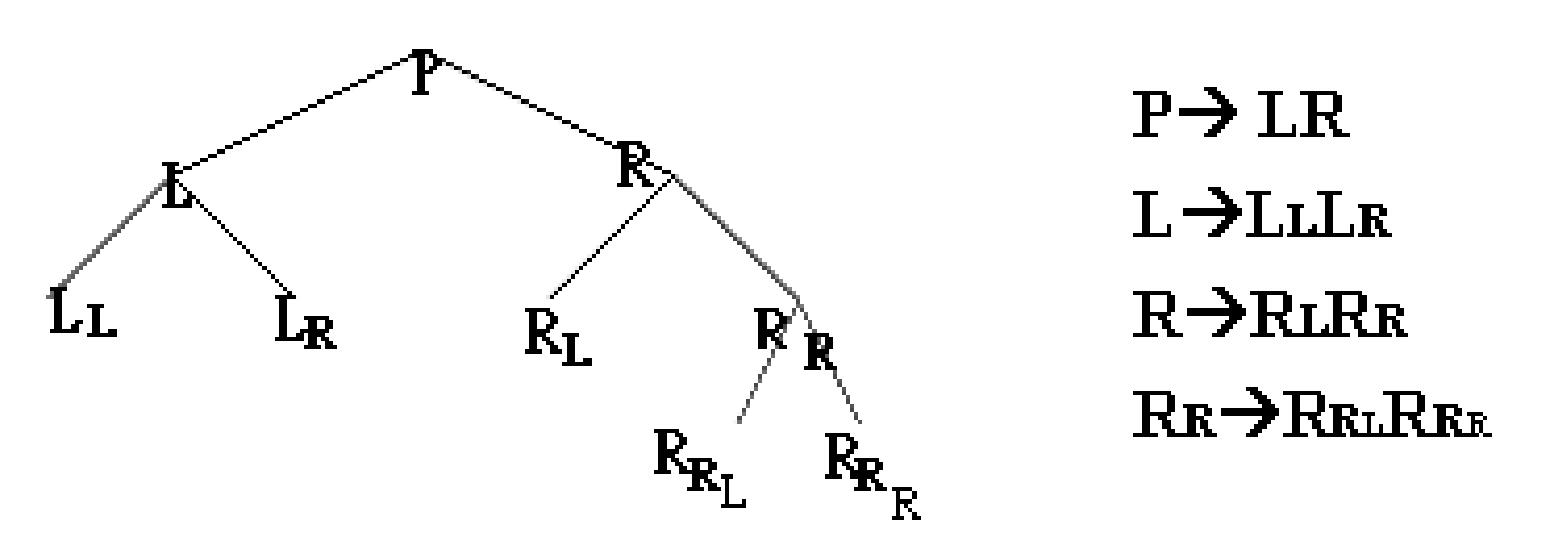}{\special{language
"Scientific Word";type "GRAPHIC";maintain-aspect-ratio TRUE;display
"USEDEF";valid_file "F";width 3.8233in;height 1.3353in;depth
0pt;original-width 5.1932in;original-height 1.7902in;cropleft "0";croptop
"1";cropright "1";cropbottom "0";filename 'fig5.jpg';file-properties
"XNPEU";}}

In Figure \ref{figure 5}, \textbf{$L_{L}$} denotes the left subtree's left
subtree, and \textbf{$R_{L}$} denotes the right subtree's left subtree. 
\textbf{$R_{R_{L}}$} and \textbf{$R_{R_{R}}$} are similar. Thus, the
rewriting rules for the tree shown in Figure \ref{figure 5} are \textbf{$%
P\rightarrow L_{L}L_{R}R_{L}R_{R_{L}}R_{R_{R}}$}.

\subsection[Bracketed strings for a rhythmic tree]{Bracketed strings for a
rhythmic tree}

\FRAME{fhFU}{3.7118in}{1.3318in}{0pt}{\Qcb{Bracketed strings for two trees.}%
}{\Qlb{figure 6}}{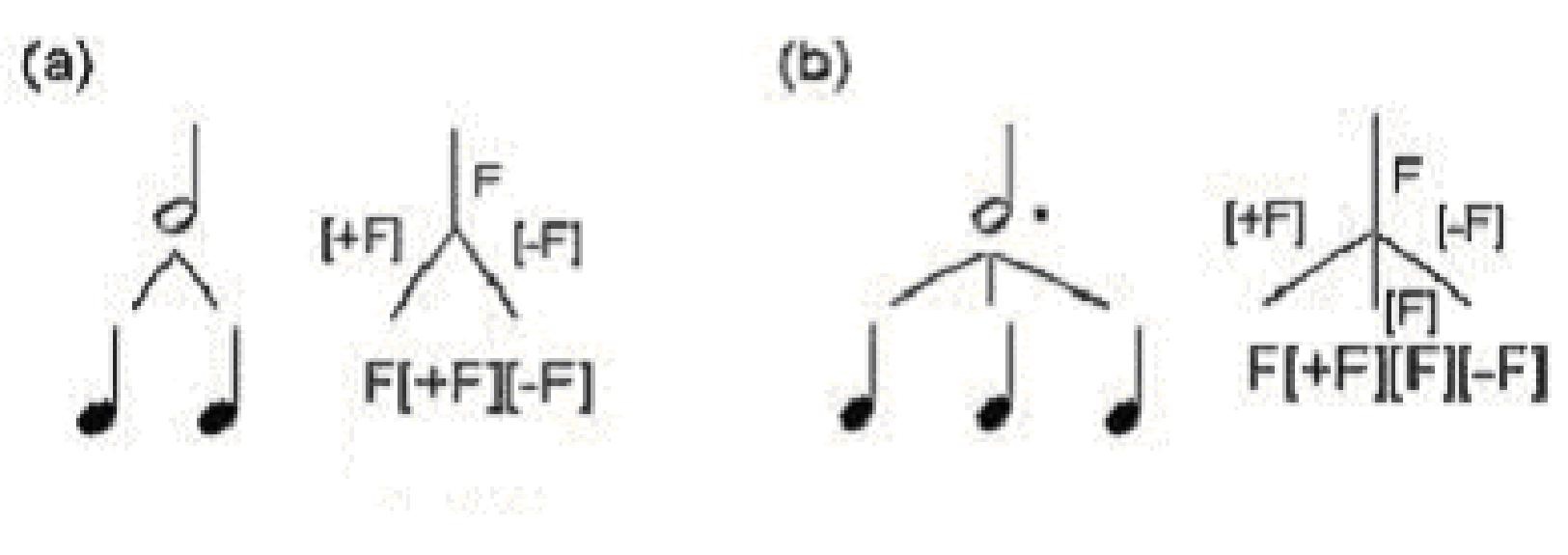}{\special{language "Scientific Word";type
"GRAPHIC";maintain-aspect-ratio TRUE;display "USEDEF";valid_file "F";width
3.7118in;height 1.3318in;depth 0pt;original-width 5.4163in;original-height
1.9164in;cropleft "0";croptop "1";cropright "1";cropbottom "0";filename
'fig6.jpg';file-properties "XNPEU";}}

As discussed above, bracketed strings can represent a hierarchical
structure, such as an axial tree. Thus, bracketed strings may provide a
suitable data structure for representing rhythm. We know that a half note
can be divided into two quarter notes, and this fact can be represented by a
tree structure, which has a parent node and two child nodes, as shown in
Figure \ref{figure 6} (a). The bracketed string of the half note and its
binary branches in Figure \ref{figure 6} (a) are F[+F][-F]: the first F is
the command for tracing the root; [+F] is the command for tracing the left
branch, and [-F] is the command for tracing the right branch. In Figure \ref%
{figure 6} (b), there is another example for a dotted half note.

\FRAME{fhFU}{2.6472in}{1.3474in}{0pt}{\Qcb{Bracketed string for Beethoven's
Piano Sonata No 6, Mov. 3.}}{\Qlb{figure 7}}{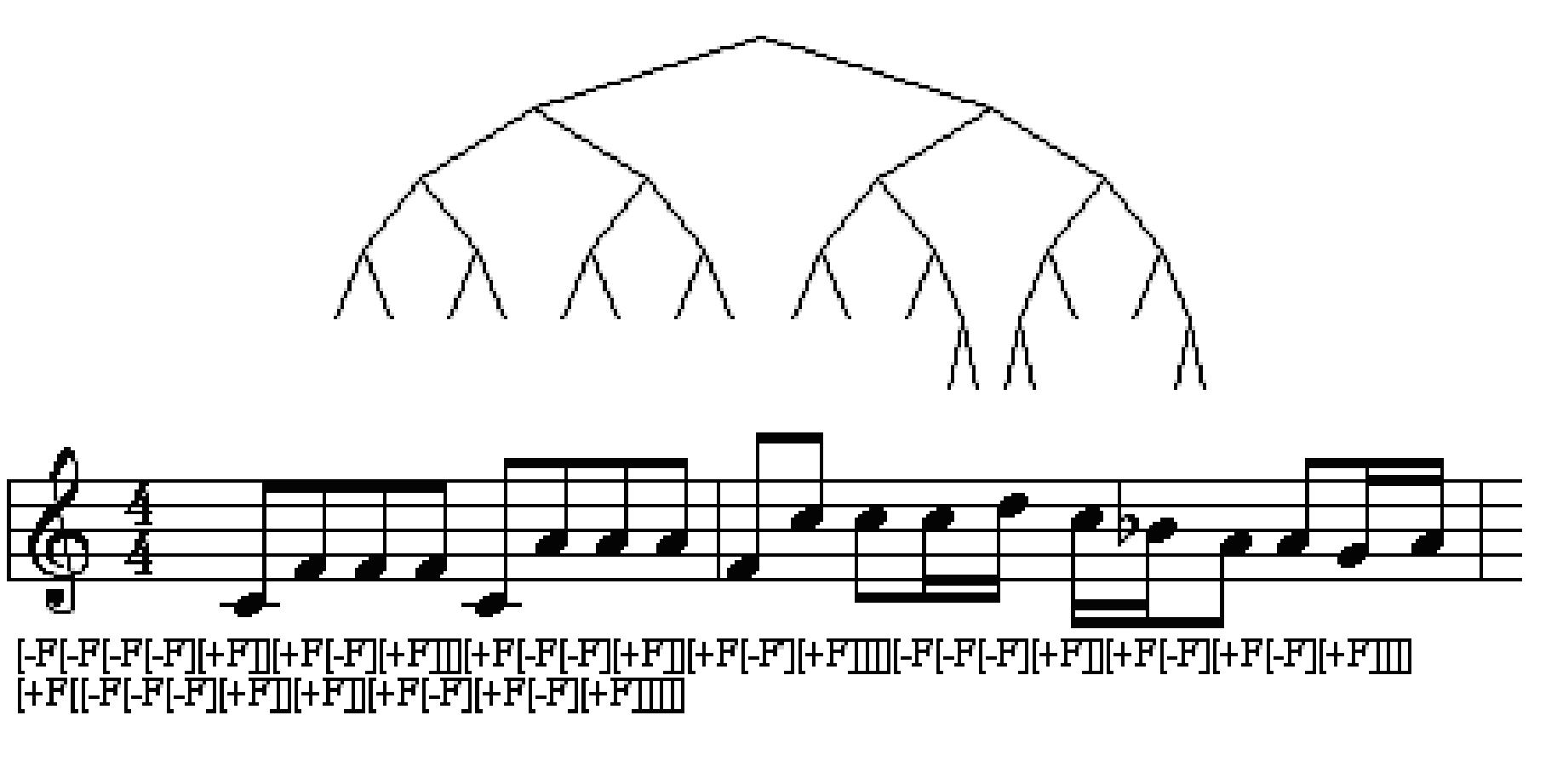}{\special{language
"Scientific Word";type "GRAPHIC";maintain-aspect-ratio TRUE;display
"USEDEF";valid_file "F";width 2.6472in;height 1.3474in;depth
0pt;original-width 6.0961in;original-height 3.0701in;cropleft "0";croptop
"1";cropright "1";cropbottom "0";filename 'fig7.jpg';file-properties
"XNPEU";}}

Since we focus here only on rhythmic trees, we can simplify the bracketed
string representations. First, our rhythmic trees have only 2 subtrees.
Second, the `F' notation for a rhythmic tree is trivial. With these two
characteristics, we may omit the `F' notation from the bracketed string and
use only four symbols, \{[, ], -, +\}, to represent rhythmic trees. In our
cases, `[\ldots ]' denotes a rhythmic (sub)tree where `\ldots ' indicates
all the bracketed strings of its subtrees. `-` indicated that the next
`[\ldots ]' notation for a tree is a left subtree of the current (sub)tree,
and `+' indicates that the next `[\ldots ]' notation is a right subtree. In
Figure \ref{figure 8}, we list the simplified rules for the subtrees shown
in Figure \ref{figure 7}.

\FRAME{fhFU}{3.2802in}{1.3292in}{0pt}{\Qcb{Bracketed strings for each node
of the rhythmic tree shown in Figure \protect\ref{figure 7}.}}{\Qlb{figure 8}%
}{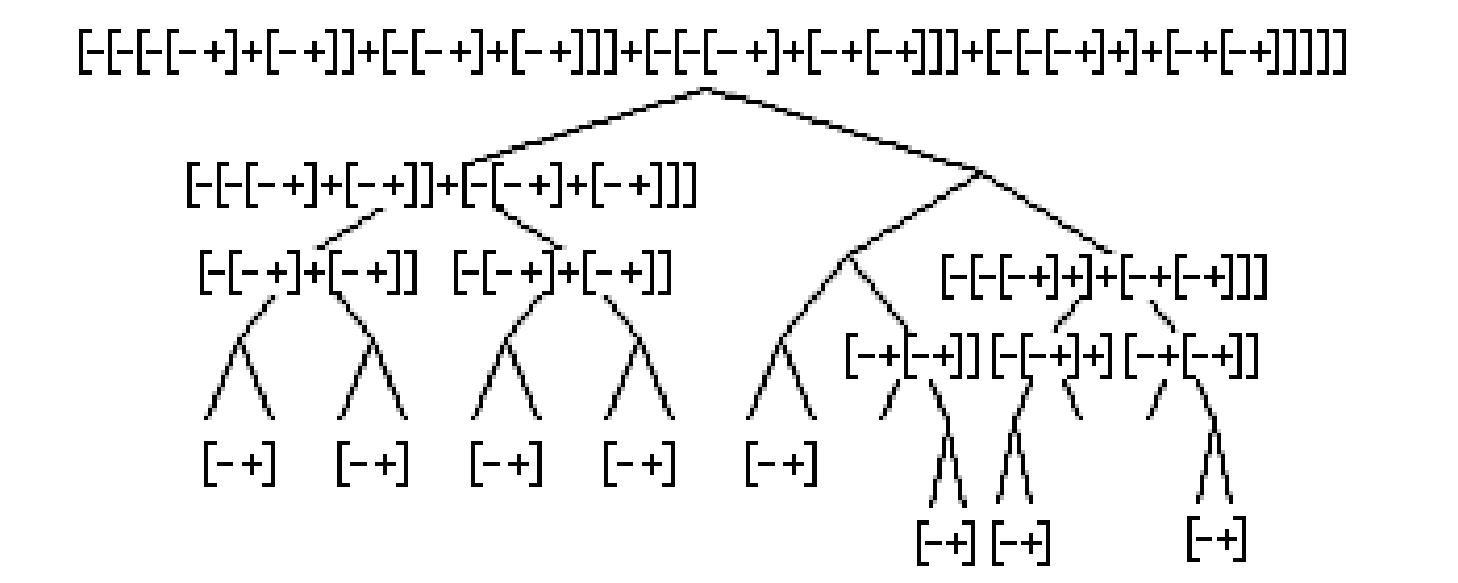}{\special{language "Scientific Word";type
"GRAPHIC";maintain-aspect-ratio TRUE;display "USEDEF";valid_file "F";width
3.2802in;height 1.3292in;depth 0pt;original-width 4.8602in;original-height
1.9432in;cropleft "0";croptop "1";cropright "1";cropbottom "0";filename
'fig8.jpg';file-properties "XNPEU";}}

Note that this rhythmic tree has no middle subtree in each node. In our
model, we always choose to divide a metrical unit into exactly two shorter
metrical subunits. We will assume that all the rhythmic trees we generate
are such binary trees. To simplify the snalysis, we also assume that the
notes are monophonic. Another example in Figure\ \ref{figure 9} is a
bracketed string for the tree of Rachmaninoff's Piano Concerto, as shown in
Figure \ref{figure 3}. Here we list all symbols in the bracketed string.

\FRAME{fhFU}{4.7469in}{1.0404in}{0pt}{\Qcb{Bracketed string for
Rachmaninoff's Piano Concerto No.3, Mov.1.}}{\Qlb{figure 9}}{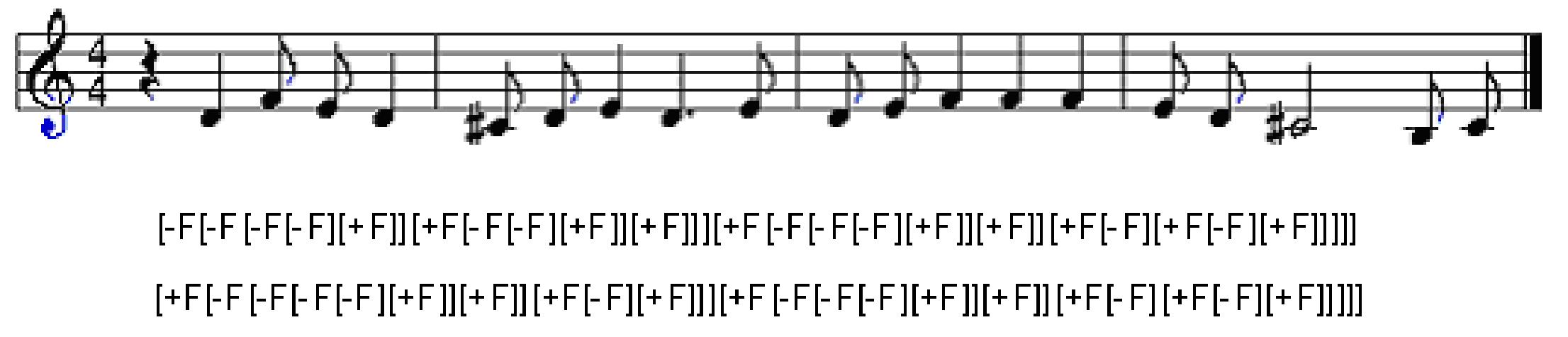}{%
\special{language "Scientific Word";type "GRAPHIC";maintain-aspect-ratio
TRUE;display "USEDEF";valid_file "F";width 4.7469in;height 1.0404in;depth
0pt;original-width 7.3734in;original-height 1.5835in;cropleft "0";croptop
"1";cropright "1";cropbottom "0";filename 'fig9.jpg';file-properties
"XNPEU";}}

\subsection[Rewriting rules for bracketed strings]{Rewriting rules for
bracketed strings}

Now, we know how to use rewriting rules to represent a tree, and we also
know how to represent a tree with a bracketed string. We can also use
rewriting rules to generate bracketed strings. In rewriting rules for
rhythmic trees, we write $\mathbf{P\rightarrow LR}$ for a tree having left
and right subtrees. Note that we call $\mathbf{L}$ and $\mathbf{R}$ the
nonterminals. Such a tree will have a bracketed string as follows:
[[-F\ldots ][+F\ldots ]]. It is clear that `[-F\ldots ]' represents the left
subtree, and that `[+F\ldots ]' represents the right subtree. Therefore, we
can replace the rewriting rules with

$\ \ \ \ \ \ \ \ \ \ \ \ \ \ \ \ \ \ \ \ \ \ \ \ \ \ \ \ \ \ \ \ \ \ \ \ \ \
\ 
\begin{array}{ccc}
P & \rightarrow & [-FL][+FR] \\ 
L & \rightarrow & .... \\ 
R & \rightarrow & ....%
\end{array}%
,$

where `\ldots ' is the rewriting rule for the bracketed string of each
subtree. In this way, we do not have to write L for a left subtree and R for
a right subtree; the orientation is already described in the bracketed
string `-F' and `+F'. Thus, we do not have to write words such as `$%
R_{R_{L}} $', `$R_{R_{R}}$', etc. Of course, we may still use such recursive
subscript representations for rules for the sake of readability. In Figure %
\ref{figure 10}, we show the rewriting rules for the bracketed string of the
tree in Figure \ref{figure 5}.

\FRAME{fhFU}{2.4163in}{2.0435in}{0pt}{\Qcb{Rewriting rules for the bracketed
string of a rhythmic tree.}}{\Qlb{figure 10}}{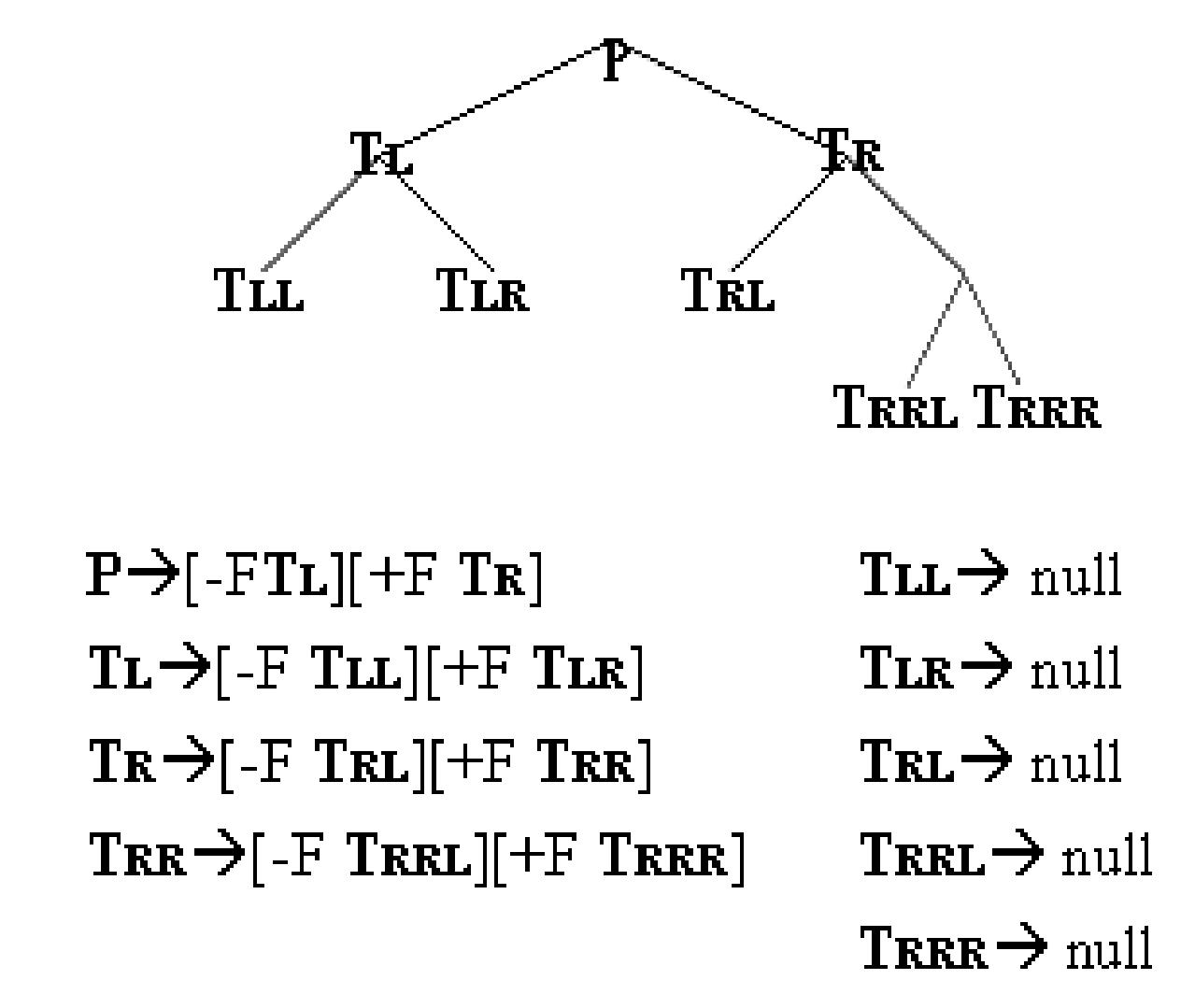}{\special{language
"Scientific Word";type "GRAPHIC";maintain-aspect-ratio TRUE;display
"USEDEF";valid_file "F";width 2.4163in;height 2.0435in;depth
0pt;original-width 4.2627in;original-height 3.5968in;cropleft "0";croptop
"1";cropright "1";cropbottom "0";filename 'fig10.jpg';file-properties
"XNPEU";}}

There are \textquotedblleft nulls\textquotedblright\ in the rules. We use
\textquotedblleft null\textquotedblright\ to represent a terminal (or a tree
node that doesn't have any child subtree). For such null-subtree rewriting
rules we simply ignore the nulls. The new rewriting rules without trivial
nulls are as follows:

$\ \ \ \ \ \ \ \ \ \ \ \ \ \ \ \ \ \ \ \ \ \ \ \ \ \ \ \ \ \ \ \ \ \ \ \ \ \ 
\begin{array}{ccc}
\mathbf{P} & \rightarrow & \mathbf{\mathrm{[-F}T_{L}\mathrm{][+F}T_{R}%
\mathrm{]}} \\ 
\mathbf{T_{L}} & \rightarrow & \mathrm{[-F][+F]} \\ 
\mathbf{T_{R}} & \rightarrow & \mathrm{[-F][+F}T_{RR}\mathrm{]} \\ 
\mathbf{T_{RR}} & \rightarrow & \mathrm{[-F][+F]}%
\end{array}%
$\textbf{$\mathrm{.}$ }

Note that there are two identical rules in the above rewriting rules: 
\textbf{$T_{L}$} and \textbf{$T_{RR}$}. This redundancy raises a question:
Can we combine them to simplify these rules? Doing so will not harm the
whole structure if the redundant rules contain only null subtrees. We will
show in the following what will happen if the rules do not contain only null
subtrees. Assume that we have the following rules:

$\ \ \ \ \ \ \ \ \ \ \ \ \ \ \ \ \ \ \ \ \ \ \ \ \ \ \ \ \ \ \ 
\begin{array}{ccc}
\mathbf{P} & \rightarrow & \mathbf{\mathrm{[-F}T_{L}\mathrm{][+F}T_{R}%
\mathrm{]}} \\ 
\mathbf{T_{L}} & \rightarrow & \mathbf{\mathrm{[-F][+F]}} \\ 
\mathbf{T_{R}} & \rightarrow & \mathbf{\mathrm{[-F][+F}T_{RR}\mathrm{]}} \\ 
\mathbf{T_{RR}} & \rightarrow & \mathbf{\mathrm{[-F][+F}T_{RRR}\mathrm{]}}
\\ 
\mathbf{T_{RRR}} & \rightarrow & \mathbf{\mathrm{[-F][+F]}}%
\end{array}%
$.\textbf{\ }

These rules can generate exactly one bracketed string and, thus, exactly one
rhythmic tree. All these rules form a rule set, which represents a unique
rhythmic tree. It is clear that \textbf{$T_{R}$} and \textbf{$T_{RR}$} are
almost the same. The only difference is that one of the subtrees is \textbf{$%
T_{RRR}$}, and that the other is \textbf{$T_{RR}$}. But they have the same
structure: they both have a right subtree and do not have a left subtree. We
can use this similarity to explore the characteristics of a music work, and
from a composer's works, we can explore his or her characteristics. We will
define two terms to express the similarity between two rewriting rules.

\section[Homomorphism and isomorphism of rewriting rules]{Homomorphism and
isomorphism of rewriting rules}

We will now study some characteristics of rewriting rules to extract similar
sections of trees. We first define the similarity between two sections as
follows:

\textbf{DEFINE : Homomorphism in rewriting rules.}

Rewriting rule $R_{1}$ and rewriting rule $R_{2}$ are homomorphic if and
only if they have the same structure. Their corresponding rhythmic trees
both have subtrees in corresponding positions or both not. That is, ignoring
all nonterminals, if rule $R_{1}$ and rule $R_{2}$ generate the same
bracketed string, then they are homomorphic by definition.

\textbf{DEFINE : Isomorphism on level $X$ in rewriting rules.}

Rewriting rule $R_{1}$ and rewriting rule $R_{2}$ are isomorphic on depth $X$
if they are homomorphic and their nonterminals are relatively isomorphic on
depth $X-1$. Isomorphic on level $0$ indicates homomorphism.

\FRAME{fhFU}{3.9219in}{1.5437in}{0pt}{\Qcb{Rhythmic tree.}}{\Qlb{figure 11}}{%
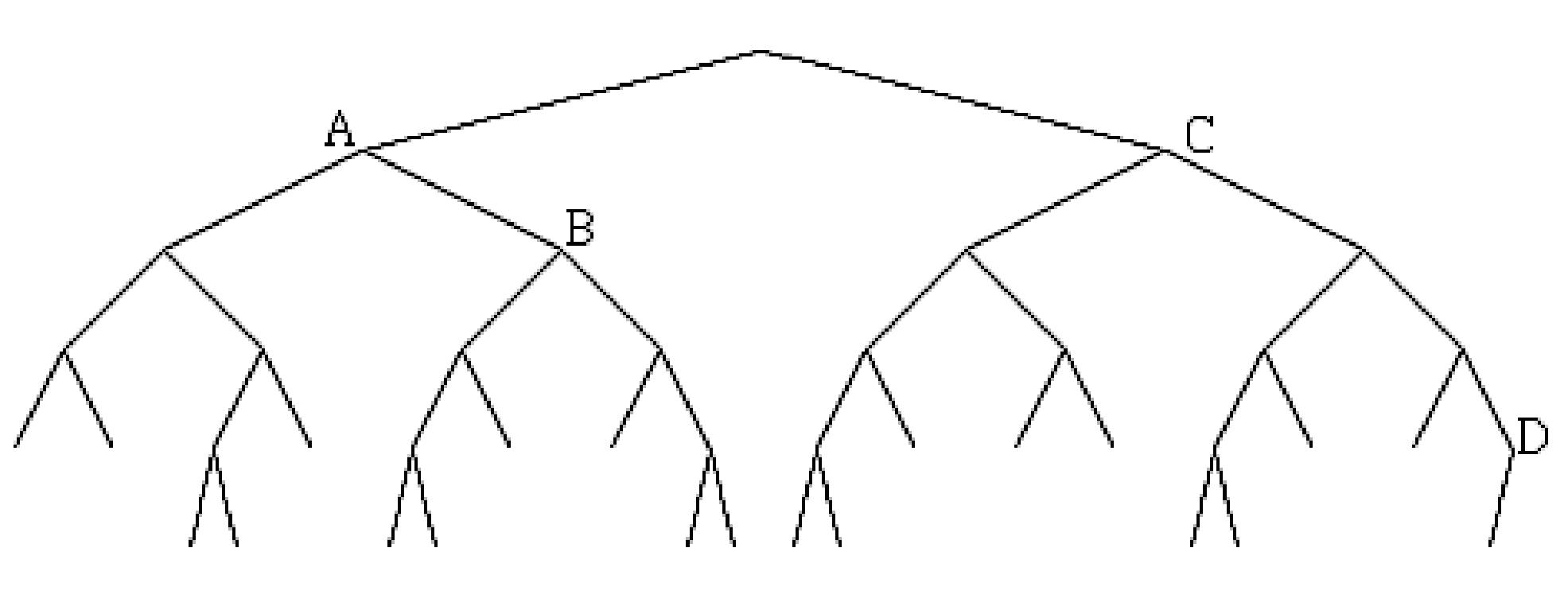}{\special{language "Scientific Word";type
"GRAPHIC";maintain-aspect-ratio TRUE;display "USEDEF";valid_file "F";width
3.9219in;height 1.5437in;depth 0pt;original-width 6.57in;original-height
2.5564in;cropleft "0";croptop "1";cropright "1";cropbottom "0";filename
'fig11.jpg';file-properties "XNPEU";}}

Here, we will use an artificial example to clarify these definitions. In
Figure \ref{figure 11}, we name the tree rooted at A, B, C, and D,
respectively, tree A, tree B, tree C, and tree D. Tree A is homomorphic to
tree B and tree C, but tree A is not isomorphic to tree D. Tree A is
isomorphic to tree C on depth 2, but they are not isomorphic on depth 3.
Tree B is isomorphic to tree C on depth 0 and 1, but not on depth 2. D is
not isomorphic to any other trees, nor is it homomorphic to any other trees.

By using bracketed strings, we can obtain a much clearer definition of
homomorphism. All homomorphic rewriting rules generate the same bracketed
strings when all nonterminals are ignored. Using the following rewriting
rules, we find that \textbf{$P$} and \textbf{$T_{L}$} are homomorphic
because if \textbf{$T_{L}$} and \textbf{$T_{R}$} are ignored, they generate
exactly the same bracketed string, [-F][+F]. We also find that \textbf{$P$}
and \textbf{$T_{RR}$} are homomorphic because if \textbf{$T_{RRR}$}, \textbf{%
$T_{L}$ }and \textbf{$T_{R}$} are ignored, they generate the same bracketed
string, [-F][+F]:

$\ \ \ \ \ \ \ \ \ \ \ \ \ \ \ \ \ \ \ \ \ \ \ \ \ \ \ \ \ \ \ \ 
\begin{array}{ccc}
\mathbf{P} & \rightarrow & \mathbf{\mathrm{[-F}T_{L}\mathrm{][+F}T_{R}%
\mathrm{]}} \\ 
\mathbf{T_{L}} & \rightarrow & \mathbf{\mathrm{[-F][+F]}} \\ 
\mathbf{T_{R}} & \rightarrow & \mathbf{\mathrm{[-F][+F}T_{RR}\mathrm{]}} \\ 
\mathbf{T_{RR}} & \rightarrow & \mathbf{\mathrm{[-F][+F}T_{RRR}\mathrm{]}}
\\ 
\mathbf{T_{RRR}} & \rightarrow & \mathbf{\mathrm{[-F][+F]}}%
\end{array}%
$.

In fact, in this example, all five rewriting rules are homomorphic to each
other. But if we add a sixth rule, \textbf{$T_{RRRR}\rightarrow \mathrm{[-F]}
$}, then it will not be homomorphic to any of the other rules.

Once we define the similarity between rules, we can classify all the rules
in the set into different subsets based on their similarity. Rules that
belong to a class are all isomorphic to each other on depth $X$. All the
rules' names are replaced with the names of the classes to which the
rewriting rules belong. After such name conversion, every new rewriting rule
represents more rhythmic trees than it was. These new rewriting rules set
can now generate more rhythmic trees, including the original one. Note that
the definition may include certain over-generalization cases. We show one
case in the last Section.

We know that these rules can generate the original bracketed string, but it
can actually do more than that. In fact, it can generate an infinite number
of bracketed strings. After performing classification, we obtain not only a
new rewriting rule set but also a context free grammar, which can be
converted into an automata. In this case, we say that this rhythmic tree can
be converted into an automata, which can generate many other rhythmic trees
that have similar characteristics. We list the rewriting rules in Table 1
for the previous example shown in Figures \ref{figure 3} and \ref{figure 9}.
Rule $Other\rightarrow null$ is ignored, there are 22 such rules in the tree
we list only one for simplicity. The classification of the rules is listed
in Table 2.

\begin{quote}
\textbf{%
\begin{tabular}[t]{|l|c|}
\hline
$\mathbf{P\rightarrow LR}$ & $\mathbf{R\rightarrow R}_{\mathbf{L}}\mathbf{R}%
_{\mathbf{R}}$ \\ \hline
$\mathbf{L\rightarrow L}_{\mathbf{L}}\mathbf{L}_{\mathbf{R}}$ & $\mathbf{R}_{%
\mathbf{L}}\mathbf{\rightarrow R}_{\mathbf{LL}}\mathbf{R}_{\mathbf{LR}}$ \\ 
\hline
$\mathbf{L}_{\mathbf{L}}\mathbf{\rightarrow L}_{\mathbf{LL}}\mathbf{L}_{%
\mathbf{LR}}$ & $\mathbf{R}_{\mathbf{LL}}\mathbf{\rightarrow R}_{\mathbf{LLL}%
}\mathbf{[+F]}$ \\ \hline
$\mathbf{L}_{\mathbf{LL}}\mathbf{\rightarrow \lbrack -F][+F]}$ & $\mathbf{R}%
_{\mathbf{LLL}}\mathbf{\rightarrow \lbrack -F][+F]}$ \\ \hline
$\mathbf{L}_{\mathbf{LR}}\mathbf{\rightarrow L}_{\mathbf{LRL}}\mathbf{[+F]}$
& $\mathbf{R}_{\mathbf{LR}}\mathbf{\rightarrow \lbrack -F][+F]}$ \\ \hline
$\mathbf{L}_{\mathbf{LRL}}\mathbf{\rightarrow \lbrack -F][+F]}$ & $\mathbf{R}%
_{\mathbf{R}}\mathbf{\rightarrow R}_{\mathbf{RL}}\mathbf{R}_{\mathbf{RR}}$
\\ \hline
$\mathbf{L}_{\mathbf{R}}\mathbf{\rightarrow L}_{\mathbf{RL}}\mathbf{L}_{%
\mathbf{RR}}$ & $\mathbf{R}_{\mathbf{RL}}\mathbf{\rightarrow R}_{\mathbf{RLL}%
}\mathbf{[+F]}$ \\ \hline
$\mathbf{L}_{\mathbf{RL}}\mathbf{\rightarrow L}_{\mathbf{RLL}}\mathbf{[+F]}$
& $\mathbf{R}_{\mathbf{RLL}}\mathbf{\rightarrow \lbrack -F][+F]}$ \\ \hline
$\mathbf{L}_{\mathbf{RLL}}\mathbf{\rightarrow \lbrack -F][+F]}$ & $\mathbf{R}%
_{\mathbf{RR}}\mathbf{\rightarrow \lbrack -F]R}_{\mathbf{RRR}}$ \\ \hline
$\mathbf{L}_{\mathbf{RR}}\mathbf{\rightarrow \lbrack -F]L}_{\mathbf{RRR}}$ & 
$\mathbf{R}_{\mathbf{RRR}}\mathbf{\rightarrow \lbrack -F][+F]}$ \\ \hline
$\mathbf{L}_{\mathbf{RRR}}\mathbf{\rightarrow \lbrack -F][+F]}$ & $\mathbf{%
Other\rightarrow null}$ \\ \hline
\end{tabular}%
}
\end{quote}

Table 1. Rewriting rules for the rhythmic tree in Figure 3.

\bigskip

\begin{tabular}{|l|l|l|l|l|}
\hline
Classification & Isomorphic & Isomorphic & Isomorphic & Isomorzphic \\ 
of rules & Depth \#0 & Depth \#1 & Depth \#2 & Depth \#3 \\ \hline
Class \#1 & $\mathbf{(21)C}_{\mathbf{1}}\mathbf{\rightarrow C}_{\mathbf{2}}%
\mathbf{C}_{\mathbf{2}}$ & $\mathbf{(3)C}_{\mathbf{1}}\mathbf{\rightarrow C}%
_{\mathbf{1}}\mathbf{C}_{\mathbf{1}}$ & $\mathbf{(1)C}_{\mathbf{1}}\mathbf{%
\rightarrow C}_{\mathbf{1}}\mathbf{C}_{\mathbf{1}}$ & $\mathbf{(1)C}_{%
\mathbf{1}}\mathbf{\rightarrow C}_{\mathbf{2}}\mathbf{C}_{\mathbf{3}}$ \\ 
&  & $\mathbf{(2)C}_{\mathbf{1}}\mathbf{\rightarrow C}_{\mathbf{2}}\mathbf{C}%
_{\mathbf{3}}$ & $\mathbf{(1)C}_{\mathbf{1}}\mathbf{\rightarrow C}_{\mathbf{2%
}}\mathbf{C}_{\mathbf{4}}$ &  \\ 
&  & $\mathbf{(1)C}_{\mathbf{1}}\mathbf{\rightarrow C}_{\mathbf{2}}\mathbf{C}%
_{\mathbf{4}}$ & $\mathbf{(1)C}_{\mathbf{1}}\mathbf{\rightarrow C}_{\mathbf{3%
}}\mathbf{C}_{\mathbf{4}}$ &  \\ 
&  & $\mathbf{(1)C}_{\mathbf{1}}\mathbf{\rightarrow C}_{\mathbf{4}}\mathbf{C}%
_{\mathbf{2}}$ &  &  \\ \hline
Class \#2 & \textbf{terminal} & $\mathbf{(4)C}_{\mathbf{2}}\mathbf{%
\rightarrow C}_{\mathbf{4}}\mathbf{C}_{\mathbf{5}}$ & $\mathbf{(1)C}_{%
\mathbf{2}}\mathbf{\rightarrow C}_{\mathbf{7}}\mathbf{C}_{\mathbf{5}}$ & $%
\mathbf{(1)C}_{\mathbf{2}}\mathbf{\rightarrow C}_{\mathbf{8}}\mathbf{C}_{%
\mathbf{4}}$ \\ 
& $\mathbf{(22)C}_{\mathbf{2}}\mathbf{\rightarrow null}$ &  &  &  \\ \hline
Class \#3 &  & $\mathbf{(2)C}_{\mathbf{3}}\mathbf{\rightarrow C}_{\mathbf{5}}%
\mathbf{C}_{\mathbf{4}}$ & $\mathbf{(1)C}_{\mathbf{3}}\mathbf{\rightarrow C}%
_{\mathbf{5}}\mathbf{C}_{\mathbf{7}}$ & $\mathbf{(1)C}_{\mathbf{3}}\mathbf{%
\rightarrow C}_{\mathbf{9}}\mathbf{C}_{\mathbf{4}}$ \\ \hline
Class \#4 &  & $\mathbf{(8)C}_{\mathbf{4}}\mathbf{\rightarrow C}_{\mathbf{5}}%
\mathbf{C}_{\mathbf{5}}$ & $\mathbf{(2)C}_{\mathbf{4}}\mathbf{\rightarrow C}%
_{\mathbf{5}}\mathbf{C}_{\mathbf{6}}$ & $\mathbf{(2)C}_{\mathbf{4}}\mathbf{%
\rightarrow C}_{\mathbf{5}}\mathbf{C}_{\mathbf{6}}$ \\ \hline
Class \#5 &  & \textbf{terminal} & $\mathbf{(4)C}_{\mathbf{5}}\mathbf{%
\rightarrow C}_{\mathbf{7}}\mathbf{C}_{\mathbf{8}}$ & $\mathbf{(4)C}_{%
\mathbf{5}}\mathbf{\rightarrow C}_{\mathbf{7}}\mathbf{C}_{\mathbf{10}}$ \\ 
&  & $\mathbf{(22)C}_{\mathbf{5}}\mathbf{\rightarrow null}$ &  &  \\ \hline
Class \#6 &  &  & $\mathbf{(2)C}_{\mathbf{6}}\mathbf{\rightarrow C}_{\mathbf{%
8}}\mathbf{C}_{\mathbf{7}}$ & $\mathbf{(2)C}_{\mathbf{6}}\mathbf{\rightarrow
C}_{\mathbf{10}}\mathbf{C}_{\mathbf{7}}$ \\ \hline
Class \#7 &  &  & $\mathbf{(8)C}_{\mathbf{7}}\mathbf{\rightarrow C}_{\mathbf{%
8}}\mathbf{C}_{\mathbf{8}}$ & $\mathbf{(8)C}_{\mathbf{7}}\mathbf{\rightarrow
C}_{\mathbf{10}}\mathbf{C}_{\mathbf{10}}$ \\ \hline
Class \#8 &  &  & \textbf{terminal} & $\mathbf{(1)C}_{\mathbf{8}}\mathbf{%
\rightarrow C}_{\mathbf{7}}\mathbf{C}_{\mathbf{5}}$ \\ 
&  &  & $\mathbf{(22)C}_{\mathbf{8}}\mathbf{\rightarrow null}$ &  \\ \hline
Class \#9 &  &  &  & $\mathbf{(1)C}_{\mathbf{9}}\mathbf{\rightarrow C}_{%
\mathbf{5}}\mathbf{C}_{\mathbf{7}}$ \\ \hline
Class \#10 &  &  &  & \textbf{terminal} \\ 
&  &  &  & $\mathbf{(22)C}_{\mathbf{10}}\mathbf{\rightarrow null}$ \\ \hline
\end{tabular}

\begin{quote}
\textbf{\ }Table 2: Classifying based on the similarity of rewriting rules.
\end{quote}

In Table 1, rules such as \textbf{$R_{RRR}\rightarrow \mathrm{[-F][+F]}$} , 
\textbf{$L_{RRR}\rightarrow \mathrm{[-F][+F]}$} are assigned to Class 2.
There are eight such rules before classification, so we write `\textbf{$%
(8)C_{2}\rightarrow \mathrm{[-F][+F]}$}'. Similar rules such as \textbf{$%
L_{R}\rightarrow L_{RL}L_{RR}$}, \textbf{$P\rightarrow LR$}, \textbf{$%
R\rightarrow R_{L}R_{R}$} are isomorphic on depth $0$, and there are $11$
such rules. They are assigned to Class 1. Class 3 and Class 4 are obtained
by following a similar classification procedure. Note that this section also
presents a new way to convert a context-sensitive grammar to a context-free
one.

\section[Rhythmic complexity]{Rhythmic complexity}

After we list the rewriting rules for a rhythmic tree and classify all those
rules, we attempt to explore the redundancy in the tree (the hidden
structure in the beats) that will be the base for building the cognitive map
[Barlow 1989]. To accomplish this, we compute the complexity of the tree
which those classified rules represent. We know that a classified rewriting
rule set is also a context free grammar, so we can define the complexity of
a rewriting rule set as follows:

\textbf{DEFINE : Topological entropy of a context free grammar.}

The topological entropy $K_{0}$ of a CFG (Context Free Grammar) can be
evaluated by means of the following three procedure [Kuich 1970; Badii and
Politi 1997]:

\ (1) For each variable $V_{i}$ with productions (in Greibach form),

\begin{center}
$V_{i}\rightarrow t_{i1}U_{i1},t_{i2}U_{i2},...,t_{ik_{i}}U_{ik_{i}},$

where $\{t_{i1},t_{i2},t_{i3}...t_{ik_{i}}\}$ are terminals and $%
\{U_{i1},U_{i2},...U_{ik_{i}}\}$ are non-terminals. The formal algebraic
expression for each variable is
\end{center}

\begin{quote}
$\ \ \ \ \ \ \ \ \ \ \ \ \ \ \ \ \ \ \ \ \ \ \ \ \ \ \ \ \ \ \ \ \ \ \ \ \ \
\ \ V_{i}=\sum_{j=1}^{k_{i}}t_{ij}U_{ij}.$
\end{quote}

\ 

(2) By replacing every terminal $t_{ij}$ with an auxiliary variable $z$, one
obtains the generating function

\begin{quote}
$\ \ \ \ \ \ \ \ \ \ \ \ \ \ \ \ \ \ \ \ \ \ \ \ \ \ \ \ \ \ \ \ \ \ \ \ \ \
\ V_{i}(z)=\sum_{n=1}^{\infty }N_{i}(n)z^{n},$
\end{quote}

where $N_{i}(n)$ is the number of words of length $n$ descending from $V_{i}$%
.

\ 

(3) Let $N(n)$ be the largest one of $N_{i}(n)$, $N(n)=max\{N_{i}(n),$over
all $i\}$. The above summation series converges when $z<R=e^{-K_{0}}$. The
topological entropy is given by the radius of convergence $R$ as

\begin{quote}
\ $\ \ \ \ \ \ \ \ \ \ \ \ \ \ \ \ \ \ \ \ \ \ \ \ \ \ \ \ \ \ \ \ \ \ \ \ \
\ \ \ K_{0}=-\ln R$.
\end{quote}

However, we have found that this definition is slightly inconvenient for our
binary tree case. Thus, we rewrite it as follows:\ 

\textbf{DEFINE : Generating function of a context free grammar. }

Assume that there are $n$ classes of rules and that each class $C_{i}$
contains $n_{i}$ rules. Let $V_{i}\in
\{C_{1},C_{2},C_{3},...,C_{n}\},U_{ij}\in \{R_{ij},i=1\symbol{126}~n,j=1%
\symbol{126}~n_{i}\}$, and $a_{ijk}\in \{x:x=1\symbol{126}~n\}$, where each $%
U_{ij}$ has the following form:

$\ \ \ \ \ \ \ \ \ \ \ \ \ \ \ \ \ \ \ \ \ \ \ \ \ \ \ \ \ \ \ \ \ \ \ \ \ 
\begin{array}{ccc}
U_{i1} & \rightarrow & V_{a_{i11}}V_{a_{i12}} \\ 
U_{i2} & \rightarrow & V_{a_{i21}}V_{a_{i22}} \\ 
... & \rightarrow & .... \\ 
U_{in_{i}} & \rightarrow & V_{a_{in_{i}1}}V_{a_{in_{i}2}}%
\end{array}%
.$

The generating function of $V_{i}$, $V_{i}(z)$ , has a new form as follows:

\begin{quote}
$\ \ \ \ \ \ \ \ \ \ \ \ \ \ \ \ \ \ \ \ \ \ \ \ \ \ \ \ \ \ \ \ \ \ \ \
V_{i}(z)=\frac{(\sum_{p=1}^{n_{i}}n_{ip}zV_{a_{ip1}}(z)V_{a_{ip2}}(z))}{%
\sum_{q=1}^{n_{i}}n_{iq}}.$
\end{quote}

If $V_{i}$ doesn't have any non-terminal, we set $V_{i}(z)=1$. With this
function, we can define the complexity of the rhythmic tree below.

\ 

\textbf{DEFINE : Complexity of rhythmic tree [6].}

After formulate the generating function $V_{i}(z)$, we intend to find the
largest value of $z$, $z^{max}$, at which $V_{1}(z^{max})$ converges. Note
that we use $V_{1}$ to denote {the rule for the root node of }the rhythmic
tree. After obtaining the largest value, $z^{max}$, of $V_{1}(z)$, we set $%
R=z^{max}$, the radius of convergence of $V_{1}(z)$. We define the
complexity of the rhythmic tree as $K_{0}=-\ln R$.

\ We use the simple example in Tables 1 and 2 (or Figure \ref{figure 3}) to
show the computation procedure of the complexity. According to our
definition the given values for the class parameters are $%
\{n=5,n_{1}=4,n_{2}=1,n_{3}=1,n_{4}=1,n_{5}=1,n_{11}=3,n_{12}=2,n_{13}=1,$ $%
n_{14}=1,n_{21}=4,n_{31}=2,n_{41}=8,n_{51}=22,a_{111}=1,a_{112}=1,a_{121}=2,$
$%
a_{122}=3,a_{131}=2,a_{132}=4,a_{141}=4,a_{142}=2,a_{211}=4,a_{212}=5,a_{311}=5 
$ $,a_{312}=4,a_{411}=5,a_{412}=5,a_{511}=2,a_{512}=3\}$.

Substituting these values in the equation, we have $V_{5}(z^{\prime })=1$
and $V_{4}(z^{\prime })=z^{\prime }$ directly. Then we obtain the formulas
for $V_{3}(z)$, $V_{2}(z)$, and $V_{1}(z)$ successively. They are

\begin{quote}
$V_{3}(z^{\prime })=\frac{(\sum_{p=1}^{n_{3}}n_{3p}z^{\prime
}V_{a_{3p1}}(z^{\prime })V_{a_{3p2}}(z^{\prime }))}{\sum_{q=1}^{n_{3}}n_{3q}}%
=\frac{z^{\prime }\times (2\times V_{5}(z^{\prime })\times V_{4}(z^{\prime
}))}{2}=z^{\prime }{}^{2}$

$V_{2}(z^{\prime })=\frac{(\sum_{p=1}^{n_{2}}n_{2p}z^{\prime
}V_{a_{2p1}}(z^{\prime })V_{a_{2p2}}(z^{\prime }))}{\sum_{q=1}^{n_{2}}n_{2q}}%
=\frac{z^{\prime }\times (4\times V_{4}(z^{\prime })\times V_{5}(z^{\prime
}))}{4}=z^{\prime }{}^{2}$

$V_{1}(z^{\prime })=\frac{(\sum_{p=1}^{n_{1}}n_{1p}z^{\prime
}V_{a_{1p1}}(z^{\prime })V_{a_{1p2}}(z^{\prime }))}{\sum_{q=1}^{n_{1}}n_{1q}}
$

$=\frac{z^{\prime }\times (3\times V_{1}(z^{\prime })^{2}+2\times
V_{2}(z^{\prime })\times V_{3}(z^{\prime })+1\times V_{2}(z^{\prime })\times
V_{4}(z^{\prime })+1\times V_{4}(z^{\prime })\times V_{2}(z^{\prime }))}{7}$

$=\frac{3z^{\prime }\times V_{1}(z^{\prime })^{2}+2\times (z^{\prime
})^{5}+2\times (z^{\prime })^{4}}{7}$
\end{quote}

Rearranging the above equation for $V_1(z)$, we obtain a quadratic equation
for $V_1(z^{\prime})$

\begin{quote}
$\ \ \ \ \ \ \ \ \ \ \ \ \ \ \ \ \ \ \ \ \ \ \ \ \ \ \ \ \ \ \ \ \ \frac{3}{7%
}(z^{\prime })V_{1}(z^{\prime })^{2}-V_{1}(z^{\prime })+\frac{2}{7}%
((z^{\prime }){}^{5}+(z^{\prime }){}^{4})=0$
\end{quote}

Solving $V_{1}(z^{\prime })$, we obtain the formula

\begin{quote}
$\ \ \ \ \ \ \ \ \ \ \ \ \ \ \ \ \ \ \ \ \ \ \ \ \ \ \ \ \ \ \ \ \ \ \
V_{1}(z^{\prime })=\frac{1\pm \sqrt{1-\frac{24}{49}((z^{\prime
}){}^{5}+(z^{\prime }){}^{4})}}{(6z^{\prime })/7}$.
\end{quote}

The radius of convergence, $R$, and complexity, $K_{0}=-\ln R$, can be
obtained from this formula.

\section[Implementation and Examples]{Implementation and Examples}

In order to compute the complexity of a rhythmic tree, we have to determine $%
R$, the radius of convergence of the rhythmic tree's rewriting rule set. We
devise strategy to judge whether the function $V_{1}(z^{\prime })$ is
convergent or divergent for a given value of $z^{\prime }$. We construct an
iteration technique to compute the value of this generating function. To
facilitate the computation, we rewrite the generating function as follows:

\begin{quote}
$V_i^m(z^{\prime}) = \frac{\sum_{p=1}^{n_i}n_{ip}z^{%
\prime}V_{a_{ip1}}^{m-1}(z^{\prime})V_{a_{ip2}}^{m-1}(z^{\prime})}{%
\sum_{q=1}^{n_i}n_{iq}}$ and

$V_i^0(z^{\prime}) = 1$.
\end{quote}

Here we use superscript $m$ in the variable $V_{i}^{m}(z^{\prime })$ to
represent the iteration count. Starting with $V_{i}^{0}(z^{\prime })$ in
each iteration, we calculate a new value, $V_{i}^{1}(z^{\prime })$. Then we
calculate $V_{i}^{2}(z^{\prime })$, $V_{i}^{3}(z^{\prime })$, ... , and $%
V_{i}^{m}(z^{\prime })$ successively, where $m$ is some positive integer
number. When $V_{i}^{m-1}(z^{\prime })$ is equal to $V_{i}^{m}(z^{\prime })$
for all rules, this means that $V_{i}^{m}(z^{\prime })$ cannot be improved
anymore, we reach convergence. Therefore, $z^{\prime }$, the number we want
to judge, is not the radius of convergence for the rules set but is smaller
than the radius of convergence. In our simulations, we set $m=1000$. This
means that if $V_{i}^{m}(z^{\prime })$ is not divergent for $m<1000$, then
we judge $z^{\prime }$ to be convergent.

Once we can judge whether $V_{i}(z^{\prime })$\ is convergent or divergent
at a number $z^{\prime }$, we can test every real number between 0 and 1 to
find the number that is right on the border of the convergent region and use
this number to calculate the radius of convergence. We may apply some
advanced techniques to search for the radius of convergence, such as binary
searching between 0 and 1. This is exactly the technique we use in our
algorithm.

Now we will present a practical example. We use Beethoven's Piano Sonatas
Nos. 1 to 32 and Mozart's Piano Sonatas Nos. 1 to 19 as an example, and show
their complexity. We list the complexity of each piano sonata by Mozart in
Figures \ref{figure12}-\ref{figure13}. In these figures, we use two
different isomorphic depths, 1 and 3, to compute the complexity. From the
figures we can see that the complexity is high for both composers. When we
use higher depth isomorphism to classify rules, the complexity will
decrease. This is because when we use higher depth isomorphism, redundancy
between rules will decrease so the complexity will also decrease. Eventually
the complexity will decrease to zero for the highest depth isomorphism.
Conversely, lower depth isomorphism brings more rules in a class; redundancy
between rules will increase and the number of classes will decrease. If the
depth of isomorphism is too low, the rules set will become too simple, thus
the complexity will also become lower. We may compute the complexity for
different depths to see the differences.

Beethoven and Mozart's work have similar complexity, but Beethoven's is
slightly higher than Mozart's. Both their complexity isomorphic on level 2
are the highest. When we use the high level isomorphism to classify rules,
the complexity of rules will decrease. Reversely, the low level isomorphism
collects many rules in a class; redundancy between rules will increase and
the number of classes will decrease. If the level of isomorphism is too low,
the rules set will become too simple, thus the complexity will also become
lower. We can try different level to see its complexity, and pick up the
level with highest complexity.

We tested a well-known music work studied by Rauscher et al. [1993]. Almost
all the previous studies on the Mozart Effect have focused on a single piece
of music, the Sonata for Two Pianos in D Major (K448). We have computed its
complexity and found that it is generally higher than that of other sonatas
by Mozart, see Figures \ref{figure12}-\ref{figure13}.

\section[Summary]{Discussion}

We have constructed the complexity for the L-system. This complexity
resembles, in some sense, the redundancy [Pollack 1990; Large et al. 1995;
Chalmers 1990]. This complexity can facilitate many other studies such as
bio-morphology, DNA analysis, gene analysis and tree similarity.

We closely followed the ideas of Barlow [Barlow 1989] and Feldman [Feldman
2000] to design this model. In his work, Barlow wrote that:
\textquotedblleft Words are to the elements of our sensations like logical
functions to the variables that compose them. We cannot of course suppose
that an animal can form an association with any arbitrary logical function
of its sensory messages, but they have capacities that tend in that
direction, and it is these capacities that the kind of representative
schemes considered here might be able to mimic.\textquotedblright\ \ Human
perception sometimes bases on external world's information redundancy. If we
can extract any rules or patterns from a certain object as part of our
cognition map for that object, it will be easy to memorize or comprehend it.
In our model, rhythms resemble the words; trees resemble the logical
functions; classes resembles rules and patterns; complexity resembles
redundancy.

Man is not inherently musical, the distinguished scientist Newton claimed;
natural singing is the sole property of birds. In contrast to our feathered
friends, humans perform and understand only what they taught ..\ldots\ This
is why humans listen to music by training. One needs such redundancy to
comprehend the music words.

But how can we pinpoint the rules or patterns in a music work, or even in a
simple rhythm that may be formless? As an attempt, we have defined
homomorphism and isomorphism so as to characterize the similarity between
sections of different rhythmic trees. But there still exist questions about
the psychological implications of these characteristics, such as the depth
of isomorphism. The proposed model can enable us to measure the
psychological complexity [Feldman 2000] of rhythms. In our studies, we have
found that different depths of isomorphism produce varying degree of
complexity. If a rhythm is very simple, its complexity will be 0. The same
situation also occurred when we used isomorphism with a very high depth
value to compute the complexity of Mozart's and Beethoven's piano sonatas.
In general, the results confirm our intuition about these musical rhythms.

Note that not all music can be properly approached with binary structures.
If the structure of a music piece is ternary, we expect that the computed
complexity will be higher than it is in reality.

We define the similarity between tree structures in Section 3. Finding
similarity between rules and classifying them in different subsets are in
some sense similar to fractal compression, see the website in Reference.
This could be an alternate way to configure rhythmic complexity. We are
still working on this. We are also working on an extension of the model to
incorporate the rhythmic complexity for polyphonic music, superposition of
different rhythms, tempo variation, grace notes, supra and irregular
subdivisions of the beat (e.g. triplets, quintuplets,...).

\FRAME{fhFU}{2.4068in}{2.1672in}{0pt}{\Qcb{Mozart's 19 Piano Sonatas, using
isomorphic depth 1.}}{\Qlb{figure12}}{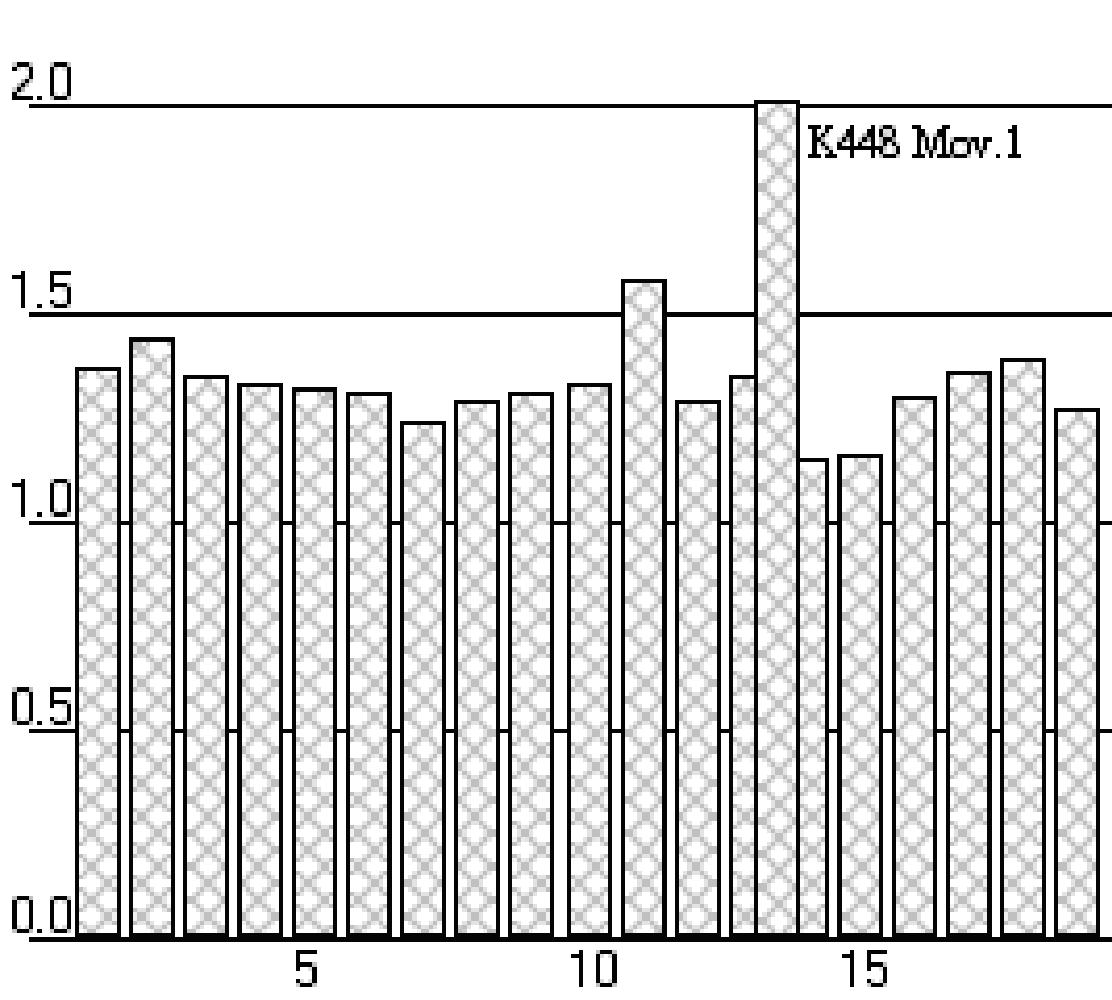}{\special{language
"Scientific Word";type "GRAPHIC";maintain-aspect-ratio TRUE;display
"USEDEF";valid_file "F";width 2.4068in;height 2.1672in;depth
0pt;original-width 3.7066in;original-height 3.333in;cropleft "0";croptop
"1";cropright "1";cropbottom "0";filename 'fig12.jpg';file-properties
"XNPEU";}}

\FRAME{fhFU}{2.4068in}{2.5045in}{0pt}{\Qcb{Mozart's 19 Piano Sonatas, using
isomorphic depth 3.}}{\Qlb{figure13}}{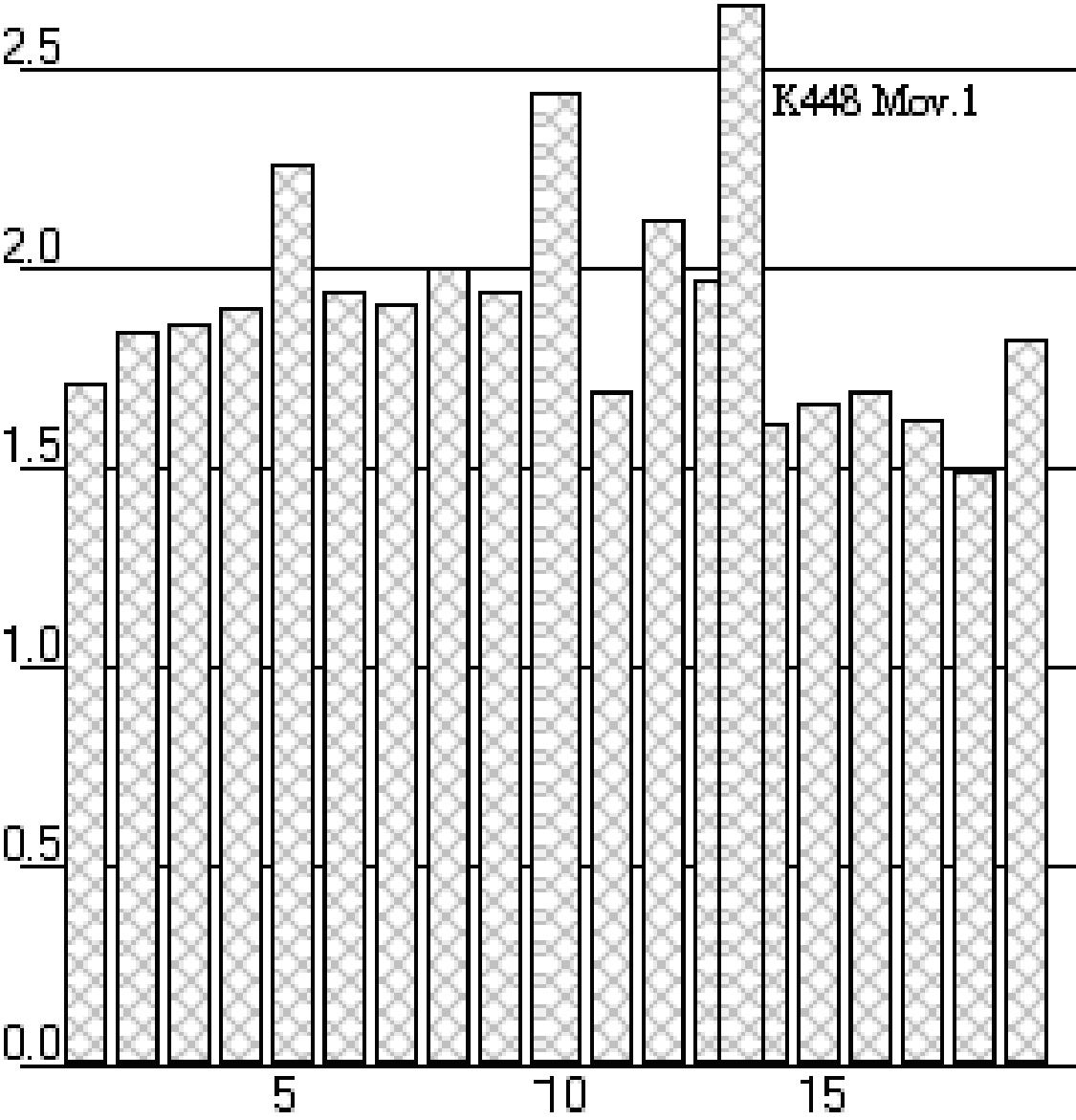}{\special{language
"Scientific Word";type "GRAPHIC";maintain-aspect-ratio TRUE;display
"USEDEF";valid_file "F";width 2.4068in;height 2.5045in;depth
0pt;original-width 3.7498in;original-height 3.9038in;cropleft "0";croptop
"1";cropright "1";cropbottom "0";filename 'fig13.jpg';file-properties
"XNPEU";}}

\end{document}